\documentclass[sn-mathphys]{sn-jnl}

\jyear{2022}%

\begin{document}

\title[Natural Intelligence]{A Theory of Natural Intelligence}


\author*[1,2,3]{\fnm{Christoph} \spfx{von der} \sur{Malsburg}}\email{malsburg@fias.uni-frankfurt.de}

\author[2,4]{\fnm{Thilo} \sur{Stadelmann}}\email{stdm@zhaw.ch}
\equalcont{These authors contributed equally to this work.}

\author[3]{\fnm{Benjamin F.} \sur{Grewe}}\email{bgrewe@ethz.ch}
\equalcont{These authors contributed equally to this work.}

\affil*[1]{\orgname{Frankfurt Institute for Advanced Studies}, \orgaddress{\city{Frankfurt}, \country{Germany}}}

\affil[2]{\orgdiv{Centre for Artificial Intelligence}, \orgname{Zurich University of Applied Sciences}, \orgaddress{\city{Winterthur}, \country{Switzerland}}}

\affil[3]{\orgdiv{Institute of Neuroinformatics}, \orgname{University of Zurich and ETH Zurich}, \orgaddress{\city{Zurich}, \country{Switzerland}}}

\affil[4]{\orgname{European Centre for Living Technology}, \orgaddress{\city{Venice}, \country{Italy}}}


\pagebreak

\abstract{{\bf Introduction:} In contrast to current AI technology, natural intelligence -- the kind of autonomous intelligence that is realized in the brains of animals and humans to attain in their natural environment goals defined by a repertoire of innate behavioral schemata -- is far superior in terms of learning speed, generalization capabilities, autonomy and creativity.  How are these strengths, by what means are ideas and imagination produced in natural neural networks?

{\bf Methods:} Reviewing the literature, we put forward the argument that both our natural environment and the brain are of low complexity, that is, require for their generation very little information and are consequently both highly structured. We further argue that the structures of brain and natural environment are closely related.

{\bf Results:} We propose that the structural regularity of the brain takes the form of net fragments (self-organized network patterns) and that these serve as the powerful inductive bias that enables the brain to learn quickly, generalize from few examples and bridge the gap between abstractly defined general goals and concrete situations.

{\bf Conclusions:} Our results have important bearings on open problems in artificial neural network research.}
  

\keywords{Ontogenesis, emergence, structural regularity, net fragments, visual perception, scene representation, homeomorphic mapping, inductive bias, autonomous behavior}

\maketitle

\section{Introduction}\label{sec1}

There may be different kinds of intelligence.
We here concentrate on the one that is epitomized in humans and animals.
This kind of intelligence is often defined
  as the ability to successfully pursue general goals in varying contexts,
  goals such as feeding oneself, avoiding danger or creating offspring.
The emphasis of our communication is on
  the neural mechanisms that generate this ability,
  our main point being that besides
  nature and nurture the process is dominated by
  a third generative factor, \emph{emergence}.
In this context, `nature' refers to the influence
  of the genes and therewith to that of evolution,
  while `nurture' to that of experience, instruction and education.
We would like to maintain here that neither quantitatively nor qualitatively
  genes and experience alone can account for
  the structure of the nervous system nor the intelligence it supports, 
  leaving a large gap to be closed by emergence.

On the quantitative side, as to `nature',
  the human genome contains one gigabyte of information (3.3 billion nucleotides of DNA \cite{Genome})
  while one petabyte is required to describe the connectivity of the human brain\footnote{$10^{14}$ synapses, 
  each taking 33 bits to address one of the $10^{10}$ neurons of the brain.}.
In the case of humans, `nurture' during the first years of life
  is provided for by an environment
  (the nursery, the family, toys, books etc.)
  that is deliberately kept simple
  and could be simulated in its visual aspects
  on the basis of a virtual reality program
  of a few gigabytes.
  Additionally, the rate at which humans absorb information
  into permanent memory is estimated \cite{Landauer}
  at only $1-2$ bits per second,
  signifying a couple of gigabits over a long lifetime.
These amounts of information are to be compared to
  the petabyte needed to list all connections in the brain.

The qualitative side is the essence of the problem we want do address:
  how can intelligence, in terms of ideas, imaginations and insights surpass so much
  everything that has been `programmed' into the genes,
  and how can it learn so fast and generalize so boldly
  beyond all the examples it has seen before?

\begin{figure}[t]
    \centering
    \includegraphics[width=1\textwidth]{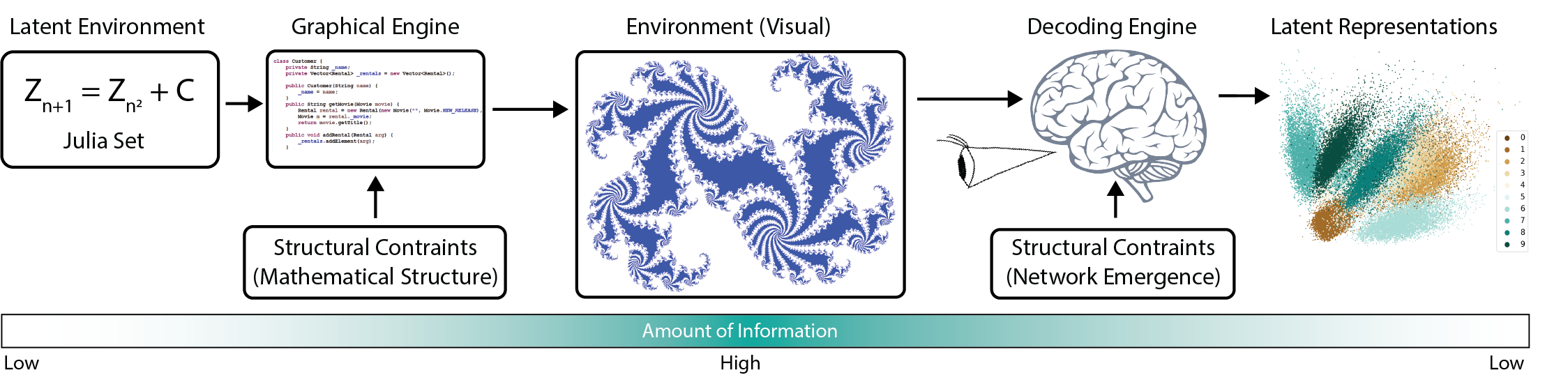}
    \caption{Illustration of Kolmogorov complexity. Julia sets (middle panel) need, literally, infinite amounts of information to be described, but very little information suffices to generate them (left; recursive definition and mathematical grammar). Perception and efficient learning are possible by reducing the flood of sensory signals produced by the environment to an underlying low-complexity description (right).}
    \label{fig:complexity}
\end{figure}

To deal with the quantitative side of the problem one has to distinguish
  the raw amount of information needed to {describe} a structure from
  the minimal amount of information required to {generate} it. 
The latter, the bit length of the shortest algorithm that can generate the structure,
  is called Kolmogorov complexity \cite{Kolmogorov} and 
  may be smaller by many orders of magnitude than the amount of information
  required to describe the structure.
An extreme example of low Kolmogorov complexity is illustrated in 
  Figure \ref{fig:complexity}.
Obviously, nature and nurture need only gigabytes to
  construct, respectively instruct, the brain.
A logical consequence of this efficiency is that
  the brain is totally dominated by structural regularity, 
  so that instead of from all randomly possible connectivity patterns among its neurons 
  nature and nurture only need to pick from a vastly smaller space of
  pre-structured patterns.
A central thesis of our communication is 
  that the \emph{structural regularity} implied by this low 
  Kolmogorov complexity \emph{acts as the domain-specific inductive bias} 
  that any system needs \cite{bias/variance,NoFreeLunch} or \cite[ch. 2.7]{mitchell1997machine} to be able to learn efficiently. 

The remainder of this paper is organized as follows: In Section \ref{sec2} we put forward
  the hypothesis that the Kolmogorov algorithm of the brain is network self-organization as studied extensively on the example of the ontogenetic development of retino-topic connections. In Section \ref{sec3} we discuss a small number of cognitive sample processes that are in need to be understood and implemented. In Section \ref{sec4} we try to make plausible how net fragments can serve as basis to solve these problems and in Section \ref{sec5} we discuss the relevance of the perspective we are creating to open problems within the current field of AI.

\section{Network Self-Organization as Kolmogorov Algorithm of the Brain}\label{sec2}

What is the type of mechanism, the concise Kolmogorov algorithm, 
  by which the connectivity of the brain and hence the structural regularity is generated under genetic guidance?
We suggest to adopt as paradigm the experimentally and theoretically well-studied
  mechanism of the ontogenesis of retinotopic connections:
The axons growing out from the retinae of vertebrates reach their 
  target structures (e.g., the optic tectum) in more or less random order,
  but after a relatively brief period they order themselves 
  so as to establish a smooth mapping conserving geometry \cite{rettecReview}.
Of all the mechanisms that have been proposed to explain the process
  only one survived comparison to experiment,
  \emph{network self-organization} \cite{ProcRoySoc, PhilTrans}.
Its general idea is quite simple.
An initial connectivity supports spontaneous activity.
This activity acts back by synaptic plasticity to alter the network,
  and this loop, from connectivity to activity and back to connectivity, continues
  until a stationary state, an {\it attractor network}, is reached.

Therefore we propose that network self-organization,
  as displayed in the retino-tectal system,
  is the Kolmogorov algorithm generating the wiring of the brain.
Sensory signals, as soon as they become available, participate in the mechanism,
  co-determining the attractor networks that are allowed to form.
Attractor networks can be characterized by optimizing two properties:
  \emph{sparsity} and \emph{consistency}.
A network is sparse if it has a small number of connections 
  converging on or diverging from any neuron and connectivity is 
  consistent if it supports high-order temporal correlations
  between sets of signals arriving at any given neuron.
This consistency means that a network is dominated by
  sets of alternative signal pathways (of approximately equal conduction delay) 
  between many pairs of source and target neurons \cite{euro1}. 
  
As result of such network self-organization, the brain develops 
  as an overlay of attractor networks (`\emph{net fragments}')
  \cite{NeuralCode}.
Each net fragment comprises a set of neurons and the connections among them.
If a set of neurons is activated again and again for a sufficient total time
  its internal connectivity can converge towards an attractor state.
There is positive feedback between the activity of the set
  and the structure of its connectivity.
As large sets of neurons are very unlikely to occur more than once,
  only small sets will be given a chance to establish themselves as net fragments.
Each neuron can be part of several net fragments. 

Many systems of low Kolmogorov complexity and implied high regularity 
  arise by emergence.
Such systems are composed of building elements that interact by 
  physical, chemical, mechanical etc. forces.
Well-known examples are soap bubbles or crystals:
Under appropriate conditions (e.g., low temperature in a liquid)
  large-scale ordered configurations arise in which
  the forces between the elements interlock such as to lend the configuration stability.
In these, weak interactive forces between the building elements (e.g., molecules)
  can achieve large-scale stability only by interlocking in consistent configurations.
 In the brain, where quite a number of connections have to conspire 
   (i.e., fire simultaneously) to activate a neuron,
   a vanishingly small subset of all possible connectivity patterns
   is singled out by their ability to dynamically self-stabilize
   as attractors of network self-organization.
   
After sufficient self-organization of the system
  larger sets of neurons can only be active 
  as interlocking net fragments,
  each of which can only become active 
  in the context of overlapping other fragments.
This favors the activation of large \emph{coherent nets},
  that is, networks which, if given sufficient time,
  would be attractors under network self-organization.
The term `net' emphasizes composition of smaller fragments,
  although a net can itself be a fragment of larger nets.

In order not to be caught in local optima,
  network self-organization needs to start from an initial state
  that already establishes a coarse global structure 
  from which it can proceed in a coarse-to-fine manner
  (for which a gradual tightening of inhibitory strength
  over the course of development \cite{Li,Lim} may be the basis). 
This initial connectivity structure, set up by earlier ontogenetic processes
  which rely on genetically controlled emergence \cite{Waddington}
  establishes gross connectivity between sensor organs, effector organs
  and the behavioral control circuits enabling animals
  to already function at the time of birth.
  
In the next sections we will give a sample of typical cognitive processes
  that are to be implemented and understood (Section \ref{sec3}), will explain
  how net fragments can serve to do so (Section \ref{sec4})
  and how this framework supports efficient learning, generalization and autonomy
  (Section \ref{sec5}).
  
  
\section{Cognitive Processes to be Implemented}\label{sec3}

What essential functions are at the basis of natural intelligence?
A lioness stalking pray in the savanna 
  has to integrate a complex array of factors
  into one coherent strategy in order to be successful.
One little disturbing factor can throw off the whole situation.
It may be that this complexity of natural situations,
  in distinction to the logical simplicity of classical AI accomplishments,
  is responsible for Moravec's paradoxon 
  (``it is comparatively easy to make computers exhibit adult level performance on intelligence tests or playing checkers, and difficult or impossible to give them the skills of a one-year-old when it comes to perception and mobility" \cite[p. 15]{Moravec}).
  
The organization of behavior within a given scene is based 
  on a representation of that scene in the brain.
\emph{Scene representation}, a contested concept \cite{Freeman, O'ReganNoe},
  does not imply static and complete rendering of detail
  as in a photographic image but
  is rather to be seen as an organizational framework
  putting abstract interpretations of scene lay-out and scene elements in relation to each other
  and to potential actions and emotional responses.
  This framework supports quick flashes of attention
  which materialize detailed reconstructions of narrow sectors of the scene.
Scene representations have to be built up by \emph{perception}.
Perception is difficult because sensory data are insufficient and ambiguous
  and contain in only entangled form the different factors 
  (shape, color, material, motion etc.) that make up the scene. 
Perception is therefore to be seen as an active process that constructs a model of the scene
  that uniquely explains the sensory signals and their changes
  under motion.

According to ethologists, animal and human behavior is defined and controlled by a number of drives (such as to satiate hunger or avoid danger),
  each of which is laid down under genetic guidance in a schematic form \cite{Ethology,Kilmer}.
A \emph{behavioral schema} can be activated by a sensory trigger feature, 
  executes a behavioral response, evaluates the outcome and is modified by the experience.
The basic behavioral machinery, 
  which serves a function analogous to a computer user acting through the 
  machine's operating system,
  is the fruit of evolutionary trial and error over many generations,
  and presumably is laid down in the style of business process models or Petri-nets in terms of relatively 
  few appropriately connected neurons or neural pools.
To integrate this basic machinery in a meaningful way 
  into the flow of scene representations
  is, however, a very complex affair and is the basic goal of \emph{learning}.

Even beyond the organization of behavior, there is a long tradition
 \cite{Kant,Piaget,Bartlett} or \cite[pp. 147--172]{Johnson}
  of discussing schemata as basis for understanding phenomena and define meaning. 
It therefore seems important to have a clear view how 
  concrete instances can be related to abstract schemata.

Learning takes place inside tasks that are governed by the behavioral drives.
The currently active drive decides which elements of the scene are relevant,
  focuses attention accordingly and curtails the scene representation to its needs.
The drive, as originally defined and further developed by experience,
  can be seen as an abstract scene description 
  that can serve to shape and interpret actual scenes as schema instantiations. 
This setting, \emph{a behavioral schema-interpreted scene, serves to powerfully
  constrain the learning process}.
  
How can these functions be understood and implemented on the basis of net fragments? 

\section{Net Fragments as Implementation Medium}\label{sec4} 

As we have argued, both our natural environment and our brain
  have very low Kolmogorov complexity (cf. Figure \ref{fig:complexity}).
We take computer graphics and virtual reality
  as models for the structure of our natural environment,
  and we take network self-organization, 
  as studied on the example of the ontogenesis of retinotopy,
  as the mechanism by which the connectivity of the brain arises.
We further note that for a system to efficiently learn it needs to have a strong bias towards
  its domain \cite{bias/variance, NoFreeLunch} or \cite[ch. 2.7]{mitchell1997machine}.
As the human brain indeed learns very efficiently
  we feel encouraged to propose the hypothesis
  that the connectivity structures that result from network self-organization,
  together with the neural dynamics that governs their activation
  in the establishment of scene representations (see below)
  are the inductive bias, the {\it a priori} structure (compare \cite{Kant}), that tunes the brain to the natural environment. 
  
In the remainder of this section we will discuss how net fragments
  can serve to implement structures and processes, taking vision as sample modality.

\subsection{Data Structure of Primary Visual Cortex}\label{secV1}

Primary visual cortex is populated with a collection of feature detector neurons 
  with an abundance of short-range lateral excitatory connections
  between them \cite{KandelSchwarz}. 
Sensory signals coming from a point within the retina in response to visual input
  activate a subset of the feature neurons whose receptive fields cover that point and its immediate environment.
Different local textures activate different such sets.
Within some months of early experience
  network self-organization will re-arrange the excitatory
  connections within each of these sets and with neurons in the neighborhood.
There are $100$ times more neurons in primary visual cortex compared
  to the number of axons coming out of the retina \cite{LeubaKraftsik1994},
  opening the way to sparse codes (as in \cite{ols_field}).
Visual input first briefly activates an exuberance of neurons,
  most of which will then be silenced 
  (by, e.g., balanced inhibition \cite{Vogels}) 
  leaving only the small subset of those neurons active
  that can support each other by lateral connections inside net fragments
  (for a model of this process see \cite{Wansch}).
(Membership in activated fragments is perhaps indicated by bursting 
  activity \cite{Payer,Naud}.)
As result of early visual experience
  texture patches (at the scale of the range of lateral connections)
  that dominate the statistics of the input
  will therefore become represented by net fragments.

This developing structure of the primary visual cortex
  resembles associative memory \cite{Hopfield1,Hopfield2},
  except that due to the short range of lateral connections
  it has the two-dimensional topological structure of the visual field
  and that its stored local states are defined on a statistical basis.
The local net fragments can be compared to the codebook vectors
  of some image compression algorithms \cite{compression}.
They can be considered as filters that interpret the actual visual input
  in terms of patterns previously experienced with statistical significance.
They suppress redundancy and regularize responses,
  as is important, for instance, to extract stereo depth \cite{Marr76} or motion.
The net fragments that respond to the surface of a coherent object
  overlap in terms of neurons and connections
  and thus form a coherent net,
  covering the object. 
Net fragments can thus be seen as implementation of the \emph{Gestalt laws} \cite{Gestalt},
  and the coherent nets they form as realization of the `force fields'
  that that movement is speaking of.
The coherence of a net covering the cortical region occupied by an object
  can serve as basis for figure-ground discrimination \cite{Malik}.
  
The example illustrates the power of net fragments as inductive bias.
Local texture-representing net fragments as such could be replaced
  by the higher-level feature neurons of deep learning systems.
However, due to neuron-wise overlap net fragments in distinction to those
  are exclusively activated when merging into a coherent field, a Gestalt.
Net fragments and their dynamics thus naturally render the 
  topological structure of the continuous surfaces
  that dominate our environment
  and allows them to be handled as a whole,
  as seen in the next subsection.
  
\subsection{Invariant Object Representation}\label{secInv}
  
A concrete object can appear in the visual cortex in an infinitude
  of versions differing in position, size, orientation and other factors.
In all these versions the object image gets represented, as just discussed,
  by coherent nets composed of local net fragments.
To store and later recognize the object when it appears in the retina in transformed version
  it is necessary to lay down connections that permit to construct, in response to visual input, 
  nets that represent views of the object independent of its position, orientation etc.
In the human brain these invariant representations presumably are located in infero-temporal cortex \cite{Rolls}.
There is psychophysical evidence \cite{Biederman} that for a large class of structured object types
  the visual system is able to construct such invariant representations
  out of shape primitives that are common to such objects.
We propose to \emph{see these shape primitives be represented as net fragments} 
  which have the flexibility to adapt to the shape of actual objects
  in spite of metric deformations, depth rotation and of course position within object-centered coordinates.
The identity and relative position of these shape-primitive-representing net fragments
  can then serve to identify the object type \cite{Biederman}
  and serve as basis for manipulation.

To enable such invariant responses to the position- etc. variant 
  representation of objects in the primary cortices 
  the proposal has been made \cite{CHA,Report,Arathorn}
  that there are rapidly switchable connections (`\emph{shifter circuits}')
  between the primary visual cortices and invariant representations in infero-temporal cortex 
  that can connect nets in those two areas in a structure-preserving way.
In both areas the object is represented by a two-dimensional field 
  of neighborhood-connected neurons.
A mapping between them is called structure preserving (`homeomorphic')
  if it is smooth (connecting neighbors in one field to neighbors in the other)
  and connects only neurons of the same type.

Simple versions of invariant object recognition on the basis of shifter circuits
  have been demonstrated \cite{Olshausen,Arathorn,WoWoLuMa}.
Shifter circuits are composed of net fragments
  and can be formed by network self-organization \cite{Fernandes}.
Active maps that connect variant images with their invariant representation as well as
  the movements and deformations of those maps constitute valuable information
  (as argued in the introduction of \cite{Arathorn}),
  so that, for instance, the shape of an object rotating in front of the eyes
  can be deduced from the deformation of this map.
The separation of visual object representation into
  external coordinates (`where') and internal structure (`what') 
  is an important example of the disentanglement of sensory patterns
  into the factors they contain.

The example of invariant object representation again illustrates the power of
  self-organized net fragments as inductive bias.
Different views onto the same object or surface are related by homeomorphy,
  and net fragments are a natural way to form homeomorphic mappings.
Such mappings, seen as dynamic entities, can track and model the movements
  of objects and surfaces in the environment and their relations to the eye.
They are an essential element needed to reconstruct and model in the brain
  the geometry, kinematics and dynamics of the natural environment.

It is tempting to see invariant visual object representation as a special case
  of the more general problem of representing the relationship between
  abstract schemata and instances they apply to.
Assuming that this relationship has the character of a homeomorphic mapping
  (preserving types of entities and their relations)
  it is conceivable that the ensemble of schema, instance and mapping between them
  comes to be represented by a coherent net
  composed of previously established fragments,
  just as in the example of invariant object representation.

\subsection{Net Fragments as Data Structure of the Mind}

There is a broad consensus of seeing neurons as atoms of meaning
  \cite{Quiroga}.
As such, individual neurons may refer to entities on any level of complexity,
  but in doing so they act merely as labels, while 
  beyond a low level of complexity they cannot render unambiguously 
    the specific structure of what they refer to. 
To do this requires a \emph{compositional data structure} (as convincingly argued in \cite{Pylyshyn}).
The lack of compositionality in artificial neural networks is referred to
  as the binding problem \cite{Report,Roskies}.

We here argue that net fragments are the brain's compositional 
  data structure and its solution to the binding problem.
It is illustrated by the visual representation of objects
  in both the variant and the invariant versions.
Individual feature neurons can, in response to visual input, fire stably 
  only in the context of a net fragment they are part of
  (see Subsection \ref{secV1} or \cite{Wansch}), 
  and this net fragment can do so only when overlapping with other net fragments
  (as neurons only fire as part of a net fragment they are part of), 
  so that the response to the input actually is that of a net spanning the whole object as currently pictured. 
This net is a one-time structure rendering the never-repeating way the object
  appears at any moment.
It responds holistically, as result of a collective effect \cite{Fano},
  just as the Gestalt psychologists \cite{Gestalt} would have it,
  and it still renders the Gestalt in minute detail.
A hierarchy of features of various complexity levels
  is represented by nested net fragments of different size. 

A good composite data structure has to be able to exert effect
  on the basis of its structure and be productive in the sense 
  of giving rise to analogous structures \cite{Pylyshyn}. 
Our example of invariant visual object recognition illustrates this condition.
The actual recognition takes place by the activation of a net 
  forming a homeomorphic point-to-point mapping
  between the invariant and the variant representation.
This net gets created by the activation of net fragments
  each of which connects a small region in the plane with the 
  variant representation (primary visual cortex) with
  a corresponding small region in the invariant representation
  (infero-temporal cortex).
These `maplets' are activated by homeomorphy 
  between the small regions they connect
  and they overlap such as to form a coherent global map between
  variant and invariant representations of the object, 
  as demonstrated in \cite{CHA, Report}.
Consequently it takes just one exposure to a new object type
  and formation and storage of a model thereof in the invariant domain
  to recognize that type of object independently of transformation state.
This explains the brain's ability \cite{biederman-bar} to recognize
  novel objects in altered position and pose after a single 
  brief exposure.
The representation of objects is compositional and productive,
  as requested by \cite{Pylyshyn}, in that the composite mappings 
  can serve any object and represent the position, size 
  and orientation of the variant object image,
  the invariant representation of an object can render a large number
  of variant versions thereof,
  and the net fragments in the two domains can be re-used for 
  an infinitude of different objects.

Compositionality applies also to representing cognitive structure
  in terms of submodalities
  (in vision, for instance, texture, color, motion, form, size, position etc.).
Whereas sensory signals contain submodalities in implicit form,
  specific submodality patterns can be represented separately
  within their own specialized cortical regions. 
Submodalities are basically independent of each other --
  object form, for instance, abstracting from position, size, surface texture or coloring.
Concrete mental objects can be constructed by linking them together
  with the help of maps of connections as described above,
  in a process analogous to the way computer graphics creates
  visual output by mapping different sub-modalities to each other
  and into the virtual camera.

Mental objects thus constructed are to be seen as larger net fragments 
  composed as mergers of pre-existing net fragments.
In a sufficiently pre-trained brain such nets, 
  once selected by input, are stable constructs
  that are attractors both in terms of 
  the fast dynamics of neural activation and inactivation
  and the slow dynamics of network self-organization.
Like in associative memory \cite{Hopfield1},
  active neurons are pushed by a number of simultaneously firing excitatory connections
  into a high-activity state,
  while silent neurons are reliably suppressed by converging inhibitory connections.
Such network states can be characterized as of high consistency 
  -- consistency between different signals arriving on individual neurons
    and consistency between the set of currently active neurons and and their connectivity. 
Network self-organization works on a slower time-scale 
  by performing something like a stochastic gradient descent of neural connections 
  with a cost function, at each individual neuron,
  that favors binary dynamics with 
  either a highly excited or deeply suppressed state.

\subsection{Neural Dynamics: How a Trained Brain Perceives}

Perception is difficult due to the paucity and ambiguity of sensory signals
  and because scene representations have to be spontaneously constructed
  such as to uniquely explain the sensory input.
Given the speed with which our brain routinely performs the task,
  this construction cannot be based on sequential memory search.
To this speed we offer the following explanation.
The sensory signals in their great ambiguity
  reach and alert all net fragments that are compatible with them.
Among these, some overlap and dynamically support each other 
  while others are mutually inhibitory.
Buried in this dynamics is (given, of course, sufficient previous experience)
  the comprehensive net that represents the scene.
Due to its pervasive consistency of all connections this net prevails
  in the dynamic process, establishes itself and inhibits all incompatible net fragments.
The activation of this net is due to a collective process \cite{Fano} 
  comparable to a phase transition \cite{Stanley} (like magnetization)
  instead of to sequential search.

\section{Relevance to Open Problems} \label{sec5}

Grave limitations \cite{marcus2018deep,shrestha2019review,zador2019critique,tuggener2021enough} of contemporary AI \cite{schmidhuber2015deep} have to do, first, with
  inability to generalize sufficiently beyond human-provided examples.
We trace this inability to the lack, in current systems, 
  of a sufficiently powerful inductive bias for learning.
Inductive biases are specific to application domains 
  \cite{bias/variance, NoFreeLunch, mitchell1997machine}.
We accordingly focus on what we call \emph{natural intelligence}
  which is tuned to solving general problems in our natural environment.
  
So far, we have argued that our natural environment has low Kolmogorov complexity,
  interpreting today's virtual reality systems (which have low complexity)
  as sufficiently convincing approximation to that environment.
We have further noted that the brain also is of low Kolmogorov complexity
  and have subscribed to the view that its connectivity structure 
  arises by emergence realized by network self-organization.
We have taken the brain's tremendous power to learn and generalize from scant examples
  as indication that emerging connectivity structures (net fragments)
  are the data structure of the brain and constitute its inductive bias for learning.

  
As to learning, two stages have to be distinguished:
  First, a system has to develop the toolbox that is necessary to model 
  the surrounding scene. 
Second, once it is in a position to model specific arrangements and processes 
  it can learn to relate them in finer and finer detail to 
  its set of behavioral schemata and the corresponding goals.
For brains, the first stage is partly reached in pre-natal development
  under genetic guidance, partly by sensory-motor experimentation
  by the young individual.
In the context of AI, this stage is modeled in the field of developmental robotics \cite{Lee}.

For brains, learning in the second stage is, by comparison to current AI technology,
  powerfully alleviated by two factors.
First, during scene construction in interaction with and under the influence 
  of a currently ruling behavioral schema the schema-relevant
  scene elements are labeled as such by their mapping to and from the schema.
This goes a long way towards credit assignment during the evaluation
  of the ongoing experience and suppresses irrelevant detail.
Second, the essential structure to be picked up from the current situation 
  (object, motion pattern, etc.)
  is already modeled as part of the scene representation,
  not only in concrete detail but also on more abstract levels.
It is therefore possible to tie together all essential elements of the 
  situation -- the relevant scene elements, 
  their relative arrangement, 
  their roles as defined in the behavioral schema --
  by strengthening or creating a small number of connections
  to fixate the experience.
This fixation has to happen at an appropriately abstract level
  (the ability to find this level being a subject for an appropriate kind of meta-learning), 
  so that the particular experience generalizes to analogous situations.
  
For AI systems, however, this generalization ability is still to be realized. The presented methods could therefore, if properly implemented, mitigate the above-mentioned problems of sample efficiency (including slow learning) and generalization in a principled and unified way, with the effect of leading to results that can approach common sense (compare with compartmentalised approaches in \cite{BOTVINICK2019408,doi:10.1073/pnas.1907373117,zellers-etal-2021-piglet}).

A second set of weaknesses of present AI technology revolves around low level of autonomy. 
In typical applications rather narrow goals are formulated by humans,
  application-specific data are collected
  and human-tuned architectures and hyper-parameter settings are empirically determined \cite{stadelmann2018deep}.  
This limits systems to specific applications and causes great expense,
  which is well illustrated by the enormous time and investment in terms of human effort
  necessary to develop autonomous vehicles.
True autonomy requires a complete (in some sense) set of abstract
  goals and behavioral schemata
  together with the ability to (learn to) relate these schemata
  to concrete situations.  
The difficulty of this is due to 
  the enormous distance in terms of abstraction
  between concrete scene elements and the representations of general goals.
We suggest that this distance is bridged 
  by homeomorphic relationships,
  and that these homeomorphic relationships can be found 
  with the help of composition of net fragments.
  
The superiority of human intelligence over that of animals is due
  to a very rich complement of culturally acquired schemata
  many of which are absorbed in verbal or symbolic form.
We are born with a behavioral repertoire 
  that is very similar in principle to that of a range of animal species,
  but soon new goals are acquired,
  grafted upon a small set of innate behavior patterns
  (such as wanting to please or imitate social partners) acting as  gateways.
It has been argued that higher intellectual abilities grow in the individual
  as layers of generalization by analogy,
  starting with the sensory-motor coordination structure 
  acquired early in life \cite{Lee,Lakoff}.
So far it hasn't been possible to model and artificially replicate that process.
We suggest that the missing element is a potently pre-conditioned data structure
  and that network self-organization is providing this pre-conditioning in our brain.

\section{Conclusion}\label{sec6}

A deep riddle of our existence is the question how the ideas and imaginations in our mind arise.  Super-natural influences and exotic force fields or quantum processes are widely invoked. According to our proposal mental phenomena appear like mathematical structures, which are singled out by the condition of logical consistency and seem to be there even before being discovered by mathematicians.

\section*{Acknowledgements}\label{sec7}
This work was conducted during the first author's stay as visiting professor at the UZH/ETH Institute of Neuroinformatics and the ZHAW Centre for AI, financed by UZH/ETH. The authors are grateful for the catalytic effect brought about by the Mindfire Foundation and helpful discussions with Rodney Douglas.

\bibliography{sn-bibliography}


\end{document}